\documentclass{article}

\usepackage[preprint,nonatbib]{neurips_2021}

\usepackage[utf8]{inputenc} 
\usepackage[T1]{fontenc}    
\usepackage{hyperref}       
\usepackage{url}            
\usepackage{booktabs}       
\usepackage{amsfonts}       
\usepackage{nicefrac}       
\usepackage{microtype}      
\usepackage{xcolor}         

\usepackage{epsfig,amsfonts,latexsym}
\usepackage{amsmath,amssymb,amsthm}
\usepackage{dirtytalk}

\newcommand{\bx}{\mathbf{x}}
\newcommand{\bX}{\mathbf{X}}
\newcommand{\bZ}{\mathbf{Z}}
\newcommand{\bK}{\mathbf{K}}
\newcommand{\bV}{\mathbf{V}}
\newcommand{\boK}{\overline{\mathbf{K}}}
\newcommand{\bfn}{\mathbf{f}}
\newcommand{\bz}{\mathbf{z}}
\newcommand{\bu}{\mathbf{u}}
\newcommand{\bv}{\mathbf{v}}
\newcommand{\by}{\mathbf{y}}
\newcommand{\bm}{\mathbf{m}}
\newcommand{\bS}{\mathbf{S}}
\newcommand{\expval}[2]{\mathbb{E}_{#1}\left[#2\right]}
\newcommand{\GP}{\mathcal{GP}}

\newcommand{\Ncal}{\mathcal{N}}
\newcommand{\Lcal}{\mathcal{L}}

\newcommand{\KL}{\text{KL}}

\title{Scaling Gaussian Processes with Derivative Information Using Variational Inference}
\author{Misha Padidar\textsuperscript{1}, Xinran Zhu\textsuperscript{1}

Leo Huang\textsuperscript{1}, Jacob R. Gardner\textsuperscript{2}, David Bindel\textsuperscript{1}
\\
\textsuperscript{1}Cornell University 
\\
\textsuperscript{2}University of Pennsylvania
}
\begin{document}
\maketitle
\begin{abstract}
Gaussian processes with derivative information are useful in many settings where derivative information is available, including numerous Bayesian optimization and regression tasks that arise in the natural sciences. Incorporating derivative observations, however, comes with a dominating $O(N^3D^3)$ computational cost when training on $N$ points in $D$ input dimensions. This is intractable for even moderately sized problems. While recent work has addressed this intractability in the low-$D$ setting, the high-$N$, high-$D$ setting is still unexplored and of great value, particularly as machine learning problems increasingly become high dimensional. In this paper, we introduce methods to achieve fully scalable Gaussian process regression with derivatives using variational inference. Analogous to the use of inducing values to sparsify the labels of a training set, we introduce the concept of inducing directional derivatives to sparsify the partial derivative information of a training set. This enables us to construct a variational posterior that incorporates derivative information but whose size depends neither on the full dataset size $N$ nor the full dimensionality $D$. We demonstrate the full scalability of our approach on a variety of tasks, ranging from a high dimensional stellarator fusion regression task to training graph convolutional neural networks on Pubmed using Bayesian optimization. Surprisingly, we find that our approach can improve regression performance even in settings where only label data is available.
\end{abstract}

\section{Introduction}
Gaussian processes (GPs) are a popular 
tool for probabilistic machine learning, widely used in scenarios where uncertainty quantification for regression is necessary \cite{schulam2017reliable, yi2016active, jones2017improving}. When used for Bayesian optimization (BO) \cite{mockus2012bayesian,snoek2012practical}, or in some regression settings found in the physical sciences like estimation of arterial wall stiffness, \emph{derivative information} may be available \cite{wu2017BOGradients,ulaganathan2016FSI}.
In these settings, we have not only noisy function values $y = f(\bx) + \epsilon$ but also noisy gradients $\nabla \by = \nabla_{\bx} f(\bx) + \boldsymbol{\epsilon}$ at some set of training points $\bX \in \mathbb{R}^{N\times D}$.
On paper, GPs are ideal models in these settings, because they allow for training on both labels $\by$ and gradients $\nabla \by$ in closed form.

Though analytically convenient, Gaussian process inference with derivative information scales poorly:
computing the marginal log likelihood and predictive distribution for an exact GP in this setting requires $O(N^3D^3)$ time and $O(N^2D^2)$ memory. Recent work has addressed this scalability in certain settings,
e.g.~for many training points in a low-dimensional space~\cite{eriksson2018scaling} or for few training points in a high-dimensional space~\cite{DeRoos2021}.
Despite these advances, training and making predictions for a GP with derivatives remains prohibitively expensive in regimes where both $N$ and $D$ are on the order of hundreds or even thousands.

We introduce a novel method to scale Gaussian processes with derivative information using stochastic variational approximations. We show that the expected log likelihood term of the Evidence Lower Bound (ELBO) decomposes as a sum over both training labels and individual partial derivatives. This lets us use stochastic gradient descent with minibatches comprised of arbitrary subsets of both label and derivative information. Just as variational GPs with inducing points replace the training label information with a set of learned \emph{inducing values}, we show how to sparsify the derivative information with a set of \emph{inducing directional derivatives}. The resulting algorithm requires only $O(M^3p^3)$ time per iteration of training, where $M \ll N$ and $p \ll D$.

We demonstrate the quality of our approximate model by comparing to both exact GPs with derivative information and DSKI from \cite{eriksson2018scaling} on a variety of synthetic functions and a surface reconstruction task considered by \cite{eriksson2018scaling}. We then demonstrate the full scalability of our model on a variety of tasks that are well beyond existing solutions, including training a graph convolutional neural network \cite{Kipf2017gcn} on Pubmed \cite{sen2008pubmed} with Bayesian optimization and regression on a large scale Stellarator fusion dataset with derivatives. We then additionally show that, surprisingly, our variational Gaussian process model augmented with inducing directional derivatives can achieve performance improvements in the regression setting even when no derivative information is available in the training set.

\section{Background}
In this section we review the background on Gaussian processes (GP) (Section \ref{sec:GP}), Gaussian processes with derivative observations (Section \ref{sec:GP_deriv}), and variational inference inducing point methods for training scalable Gaussian processes (Section \ref{sec:svgp}).

\paragraph{Derivative notation.} Throughout this paper for compactness we abuse notation slightly and use $\partial_{j} \by_{i}$ to refer to the $j$th element of $\nabla \by_{i}$. In this particular case, this would correspond to the partial derivative observation in dimension $j$ for training example $\bx_i$. We also use $\partial_{\bv} \by_{i}$ to refer to the directional derivative in the direction $\bv$, i.e. $\nabla \by_{i}^{\top}\bv$.

\subsection{Gaussian processes}
\label{sec:GP}

A Gaussian process (GP) is a distribution over functions $f\sim\GP(\mu(\bx),k(\bx,\bx'))$ specified by mean and covariance function $\mu,k$ \cite{rasmussen2003gaussian}. Given data points $X = \{\bx_1,...,\bx_{N} \}$ and function observations $\bfn = \{f(\bx_1),...,f(\bx_{N})\}$, placing a GP prior assumes the data is normally distributed with $\bfn \sim \Ncal(\mu_X,K_{XX})$ where $\mu_X$ is the vector of mean values at $X$ and $K_{XX} \in \mathbb{R}^{N \times N}$ is a covariance matrix. 
Conditioning on noisy observations $\by = \bfn + \epsilon$ where $\epsilon\sim\mathcal{N}(0,\sigma^2I)$
induces a posterior distribution $p(\bfn^*|\by)$ over the value of $f$ at points $\bx^*$, which is Gaussian with mean $\mu(\bx^*) - K_{\bx^*X}(K_{XX}+\sigma^2I)^{-1}(\bfn - \mu_X)$ and covariance $k(\bx^*,\bx^*) - K_{\bx^*X}(K_{XX}+\sigma^2I)^{-1}K_{X\bx^*}$. Thus, standard GP inference takes $O(N^3)$ time. Hyperparameters such as $\sigma,\theta$ are generally estimated by Maximum Likelihood. The log marginal likelihood
\begin{equation}
    \mathcal{L}(X, \theta, \sigma | \by)
    = -\frac{1}{2}(\by - \mu_X)^T(K_{XX}+\sigma^2I)^{-1}(\by- \mu_X) - \frac{1}{2}\log|K_{XX}+\sigma^2I| - \frac{n}{2}\log(2\pi)
\end{equation}
can be optimized with methods like BFGS \cite{nocedal2006numerical} at a complexity of $O(N^3)$ flops per iteration.

\subsection{Gaussian processes with derivatives}
\label{sec:GP_deriv}
GPs can leverage derivative information to enhance their predictive capabilities. Notably, as differentiation is a linear operator, the derivative of a GP is a GP \cite{parzen1999stochastic}. Derivative observations can then be naturally included in a GP by defining a multi-output GP over the tuple of function observations and partial derivative observations $(\by,\nabla \by)$ \cite{rasmussen2006GPML}. The GP has mean and covariance functions
\begin{equation}
    \mu^{\nabla}(\bx) = \begin{bmatrix}\mu(\bx) \\ \nabla_{\bx} \mu(\bx)\end{bmatrix}, \quad\quad
    k^{\nabla}(\bx,\bx') = \begin{bmatrix}k(\bx,\bx') & \big(\nabla_{\bx'} k(\bx,\bx')\big)^T \\ \nabla_\bx k(\bx,\bx') & \nabla^{2} k(\bx,\bx')\end{bmatrix}.
\end{equation}
While including partial derivative observations can enhance prediction of $f$, and vice versa, a price is paid in the computational cost, as training and inference of GPs with derivatives scale as $O(N^3D^3)$. This scalability issue has been addressed in the low $D$ setting, and is discussed in section \ref{sec:related_work}.

\subsection{Stochastic Variational Gaussian Processes}
\label{sec:svgp}

Inducing point methods \cite{snelson2006sparse,quinonero2005unifying,titsias2009variational,hensman2015scalable} achieve scalability by introducing a set of \emph{inducing points}: an \say{artificial data set} of points $\bZ= [\bz_{j}]_{j=1}^{M}$ with associated \emph{inducing values}, $\bu = [u_{j}]_{j=1}^{M}$. Stochastic Variational Gaussian Processes (SVGP) \cite{hensman2013gaussian} augment the GP prior $p(\bfn \mid \bX) \to p(\bfn \mid \bu, \bX, \bZ)p(\bu \mid \bZ)$ and then learn a variational posterior $q(\bu) = \Ncal(\bm,\bS)$. Inference for an observation $\by^*$ at $\bx^*$ takes time $O(M^3)$:
\begin{equation}
    q(\by^*) = \Ncal(y^*; K_{\bx^*Z}K_{ZZ}^{-1}\bm,\sigma_\bfn(\bx^*)^2 + \sigma^2)
\end{equation}
where $\sigma_\bfn(\bx)^2 = K_{\bx\bx} - K_{\bx Z}K_{ZZ}^{-1}K_{Z\bx} + K_{\bx Z}K_{ZZ}^{-1}SK_{ZZ}^{-1}K_{Z\bx}$ is the data-dependent variance. Using Jensen's inequality and the variational ELBO \cite{hensman2015scalable,hoffman2013stochastic}, SVGP develops a loss that is separable in the training data and amenable to stochastic gradient descent (SGD) \cite{robbins1951stochastic}, as the Kullback-Leibler (KL) divergence regularization only depends on $\bu$
\begin{equation}
    \Lcal_{\text{SVGP}} = \sum_{i=1}^N \left\{\log\Ncal(y_i|\mu_\bfn(\bx_i),\sigma^2) -\frac{ \sigma_\bfn(\bx_i)^2}{2\sigma^2}\right\} - \KL\left[ q(\bu)||p(\bu) \right].
\end{equation}
This loss is minimized over the variational parameters $\bm,\bS$ and the GP hyperparameters $\theta$. Training with SGD on mini-batches of $B$ data points brings the time per iteration to $O(BM^2 + M^3)$.

While SVGP scales well, its predictive variances are often dominated by the likelihood noise \cite{jankowiak2020parametric}. Modeling derivatives necessarily involves heteroscedastic noise, or at least different noise for the function values and gradients, which may make SVGP with a Gaussian likelihood ill-suited to the task. The Parametric Gaussian Process Regressor (PPGPR) achieves heteroscedastic modeling by using the latent function variances without modifying the likelihood by symmetrizing the dependence of the loss on the data-dependent variance $\sigma_\bfn(\bx_i)^2$ term
\begin{equation}
    \Lcal_{\text{PPGPR}} = \sum_{i=1}^N \log\Ncal(y_i|\mu_\bfn(\bx_i),\sigma^2 + \sigma_\bfn(\bx_i)^2) - \KL[q(\bu)||p(\bu)].
\end{equation}
In Section~\ref{sec:experiments}, we evaluate our approach as an extension to both SVGP and PPGPR, and find 
that PPGPR gives significant performance gains.

\section{Related Work}
\label{sec:related_work}
DSKI and DSKIP \cite{eriksson2018scaling}, derivative extensions of SKI \cite{wilson2015ski} and SKIP \cite{gardner2018skip}, are among the first methods to address scaling Gaussian processes with derivative information in a low dimensional setting. DSKI and DSKIP approximate derivative kernels by differentiating interpolation kernels $\nabla k(\bx,\bx') \approx \sum_i \nabla w_i(\bx)k(\bx_i,\bx')$ where $w_i(\bx)$ are interpolation weights used in SKI. 
Like SKI, DSKI suffers from the curse of dimensionality, and matrix-vector products cost $O(ND6^D + M\log M)$ time. DSKIP improves the dependence on $D$, but still costs $O(D^2(N + M\log M + r^3N\log D))$ to form the approximate kernel matrices, where $r\ll N$ is the effective rank of the approximation. 
Thus while these methods exhibit high model fidelity, they are limited to low dimensional settings.

Recently \cite{DeRoos2021} introduced an exact method for training GPs with derivatives in 
time $O(N^2 D  + (N^2)^3)$,
which improves on the naive $O(N^3D^3)$ when $N <D$. This method is not applicable as $N$ grows moderately large, while our paper chiefly focuses on the high-$N$ and high-$D$ setting.

Bayesian optimization with derivatives was considered in \cite{wu2017BOGradients}. Here, the authors consider conditioning on directional derivatives to achieve some level of scalability, but the dataset sizes considered never exceed $N$ of around $200$ or $D$ of around $8$. Distinct from their consideration of directional derivative information, we will be equipping \emph{each inducing point} in a sparse GP model with its own set of distinct directional derivatives, allowing the model to learn derivatives in many directions in regions of space where there are a large number of inducing points. 

\section{Methods}

Our goal is to enable training and inference on data sets with large $N$ and $D$ when derivatives are available. We will present our method in three steps. First, we describe a naive adaptation of stochastic variational Gaussian processes to the setting with derivatives. Second, we argue that this adaptation again scales poorly in $D$. Finally, we show that using additional sparsity on the derivatives gives us scalability in both $N$ and $D$.

\subsection{Variational Gaussian processes with derivatives.}
\label{sec:grad_svgp}

As described in Section \ref{sec:svgp}, SVGP creates a dataset of \emph{inducing points} $\bZ =[\bz_{j}]_{j=1}^{M}$ with labels (or \emph{inducing values}) $\bu = [u_{j}]_{j=1}^{M}$. Assume we are given a dataset $\bX =[\bx_{i}]_{i=1}^{N}$ with labels $\by = [y_{i}]_{i=1}^{N}$ and derivative observations $\nabla \by = [\nabla y_{i}]_{i=1}^{N}$. 
A natural extension of SVGP to this data is to augment the standard inducing dataset with \emph{inducing derivatives}, $\nabla \bu = [\nabla u_j]_{j=1}^{M}$, each of length $D$, so that each inducing point becomes a triple $(\bz_{j}, u_{j}, \nabla u_{j})$. This corresponds to a new augmented GP prior: 
\begin{equation}
    p(\bfn,\nabla \bfn\mid\bX)\to p(\bfn, \nabla \bfn\mid\bu, \nabla \bu, \bX, \bZ)p(\bu, \nabla \bu\mid\bZ).
\end{equation}
Analogous to SVGP, we introduce a variational posterior:
\begin{equation}
    q(\bu, \nabla \bu) = \mathcal{N}\left(\bm^{\nabla}, \bS^{\nabla}\right)= \mathcal{N}\left(\begin{bmatrix}\bm \\ \nabla \bm \end{bmatrix}, \begin{bmatrix}\bS & \nabla \bS \\ \nabla \bS^{\top} & \nabla^{2} \bS\end{bmatrix}\right).
    \label{eq:grad_svgp_posterior}
\end{equation}
Here, $\bm$ and $\nabla \bm$ are trainable parameters learned by maximizing the ELBO. We abuse notation and call the second portion of the vector $\nabla \bm$ because these variational mean parameters correspond to the $M\times D$ inducing derivative values. This also holds for the matrices $\nabla \bS$ and $\nabla^{2} \bS$.

With this augmented variational posterior, the ELBO
becomes:
\begin{equation}
    \expval{q(\bfn, \nabla \bfn)}{\log p(\by, \nabla \by \mid \bfn, \nabla \bfn)} \\- \KL(q(\bu, \nabla \bu)||p(\bu, \nabla \bu))
    \label{eq:ELBO}.
\end{equation}
Assuming the typical iid Gaussian noise likelihood for regression and expanding the first term further:
\begin{equation}
\begin{split}
    \lefteqn{\expval{q(\bfn, \nabla \bfn)}{\log p(\by, \nabla \by \mid \bfn, \nabla \bfn)} =} & \\
    &\quad \sum_{i=1}^{N} \expval{q(f_{i})}{\log p(y_{i} \mid f_{i})} + \sum_{i=1}^{N}\sum_{j=1}^{D} \expval{q(\partial_{j} f_{i})}{\log p(\partial_{j} y_{i} \mid \partial_{j} f_{i})}.
\end{split}\label{eqn:deriv_elbo}
\end{equation}
%
Here, we have used linearity of expectation and the conditional independence between $y_{i}$ and $\partial_{j} f_{i}$ given $f_{i}$ to show that the term of the ELBO that depends on training data decomposes as a sum over labels $y_{i}$ and partial derivatives $\partial_{j} y_{i}$. Thus, minibatches can contain an arbitrary subset of labels $(\bx_{i}, y_{i})$ and partial derivatives 
$(\bx_{i}, \partial_{j} y_{i})$,
and the minibatch size $B$ remains independent of $N$ and $D$.

The moments of $q(\bfn, \nabla \bfn) = \int p(\bfn, \nabla \bfn \mid \bu, \nabla \bu) p(\bu, \nabla\bu)\, d\bu \, d\nabla \bu$ are similar to those in SVGP, but the kernel matrices have been augmented with derivatives (i.e., using the kernel $k^{\nabla}(\bx, \bx')$):
\begin{equation}
    \mu_{\bfn, \nabla \bfn} = K^{\nabla}_{XZ}\big(K^{\nabla}_{ZZ}\big)^{-1}\bm^\nabla,
    \quad 
    \Sigma_{\bfn, \nabla \bfn} = 
    K^{\nabla}_{XX} + 
    K^{\nabla}_{XZ}K^{\nabla -1}_{ZZ}(\bS^\nabla 
    - K^{\nabla}_{ZZ})\big(K^{\nabla}_{ZZ}\big)^{-1}K^{\nabla}_{ZX}.
    \label{eq:var_post_mean_and_cov}
\end{equation}
Here, $\bK^\nabla_{XX}$ is a $B \times B$ matrix that corresponds to a randomly sampled subset of label and partial derivative information. $\bK^\nabla_{XZ}$ is $B \times M(D+1)$, and both $\bS^\nabla$ and $\bK^\nabla_{ZZ}$ are $M(D+1) \times M(D+1)$. Similarly, the KL divergence $\text{KL}(q(\bu, \nabla \bu) || p(\bu, \nabla \bu))$ involves multivariate Gaussians with covariance matrices of size $M(D+1) \times M(D+1)$. As a result, the running time complexity of an iteration of training under this framework is $O(M^3D^3)$ which, grows rapidly with dimension.

\subsection{Variational Gaussian processes with directional derivatives.}

The procedure above is deceptively expensive despite the asymptotic complexity of a single iteration. Because a minibatch of size $B$ contains an arbitrary subset of the $N$ labels and $ND$ partial derivatives rather than simply a subset of the $N$ labels, each epoch in the above procedure must process roughly $\frac{N+ND}{B}$ minibatches, rather than the usual $\frac{N}{B}$. Additionally, because $\bK^\nabla_{ZZ}$ is of size $M(D+1) \times M(D+1)$, the above procedure is also analogous to SVGP using $M(D+1)$ inducing points rather than using $M$. While minibatch training adapts readily to $N(D+1)$ training examples, in practice it is rare to use significantly more than $1000$ inducing points, which can require specialized numerical tools to make scale even to $M=10000$ \cite{pleiss2020CIQ}. In practice, $M(D+1)$ would rapidly result in matrices $\bK^\nabla_{ZZ}$ that make training infeasibly slow. 

To make the matrix $\bK^\nabla_{ZZ}$ not directly scale with the input dimensionality, we replace the inducing derivatives from equation \eqref{eq:grad_svgp_posterior} with \emph{inducing directional derivatives}. Rather than the triplet $(\bz_{i}, u_{i}, \nabla \bu_{i})$ with $\nabla \bu_{i}$ having dimension $D$, each inducing point is now equipped with a set of $p$ distinct directional derivatives $(\bz_{i}, u_{i}, \partial_{\bV_{i1}} u_{i}, ..., \partial_{\bV_{ip}} u_{i})$ in the directions $\bv_{i1},...,\bv_{ip}$. We include the inducing directions $\overline{\bV} = [\bV_{1} \cdots \bV_{M}] \in \mathbb{R}^{Mp \times D}$ as trainable parameters. 

\paragraph{GPs with Directional Derivatives.} Similar to how we built the derivative kernel matrix in Section \ref{sec:GP_deriv}, we may define a multi-output GP over an unknown function and its directional derivatives. For a point $\bz_{i}$ and some direction $\bv_{i}$ and another point and direction $\bz_{j}$ and $\bv_{j}$ the directional-derivative covariance function is:
\begin{align}
    k^{\partial_{v_{i}} \partial_{v_{j}}}(\bz_{i}, \bz_{j})=  \begin{bmatrix}k(\bz_{i}, \bz_{j}) & \nabla_{\bz_{j}} k(\bz_{i}, \bz_{j})^{\top} \bv_{j} \\  \bv_{i}^{\top}\nabla_{\bz_{i}} k(\bz_{i}, \bz_{j}) & 
    \bv_{i}^{\top}\nabla^{2}_{\bz_{i}\bz_{j}} K(\bz_{i}, \bz_{j})\bv_{j}\end{bmatrix},
    \label{eq:directional_kernel}
\end{align}
which 
is of size $2 \times 2$ rather than $(D+1) \times (D+1)$ as with $k^{\nabla}(\cdot, \cdot)$.

Given $Mp$ inducing directions $\overline{\bV}$, $p$ per each of the $M$ inducing points, the relevant kernel matrices (1) between all pairs of inducing values and directional derivatives, $\boK_{ZZ}$, and (2) between all inducing values and training examples with full partial derivative observations, $\boK_{XZ}$, are:
\begin{equation}
    \boK_{ZZ} = \begin{bmatrix}\bK_{ZZ} & \nabla_{Z} \bK_{ZZ} \overline{\bV} \\ \overline{\bV}^{\top} \nabla_{Z} \bK_{ZZ} & \overline{\bV}^{\top} \nabla^{2}_{ZZ} \bK_{ZZ} \overline{\bV}\end{bmatrix},
    \quad
    \boK_{XZ} = \begin{bmatrix}\bK_{XZ} & \nabla_{Z} \bK_{XZ}\overline{\bV} \\ \nabla_{X} \bK_{XZ} & \nabla^{2}_{Z} \bK_{XZ} \overline{\bV}\end{bmatrix},
\end{equation}
the first of which has shape $M(p+1)\times M(p+1)$. Constructing $\boK_{ZZ},\boK_{XZ}$ is inexpensive as we compute them directly from the directional derivative kernel \eqref{eq:directional_kernel}, rather than computing the full gradient kernel $k^\nabla$ and multiplying by the directions $\overline{\bV}$ which would incur a cost of $O(M^2D^2)$. 

Variational inference with this model is nearly identical to inference with full inducing gradients. We define a variational posterior, this time over the $M(p+1)$ inducing values and directional derivatives:
\begin{equation}
    q(\bu, \partial_{\bV} \bu) = \mathcal{N}\left(\overline{\bm}, \overline{\bS}\right)
\end{equation}
where $\overline{\bm} \in \mathbb{R}^{M(p+1)}$ and $\overline{\bS} \in \mathbb{R}^{M(p+1) \times M(p+1)}$. Inference proceeds by computing $q(\bfn, \nabla \bfn)$ from \eqref{eq:var_post_mean_and_cov} by replacing the kernel matrices $K^{\nabla}_{XZ}$ and $K^{\nabla}_{ZZ}$ with our directional derivative variants $\boK_{XZ}$ and $\boK_{ZZ}$. Because the structure of the ELBO remains unchanged,
the training labels and partial derivatives can again be subsampled to form minibatches of size $B$, 
yielding $\boK_{XZ} \in \mathbb{R}^{B \times M(p+1)}.$

\paragraph{Derivative modeling with $p \ll D$.} A key feature of this framework is that it allows for the use of a different number $p$ of directional derivatives per inducing point than the number of partial derivative observations per training point. Particularly for kernel matrices involving training examples with full partial derivative information, using $p\ll D$ directional derivatives keeps the matrix dimension small and independent of $D$. Nevertheless, allowing each inducing point to have its own set of learnable directions enables the model to learn many derivative directions where necessary in the input space by placing multiple inducing points with different directions nearby. A notable case is when each inducing point $\bz_i$ has the $p = D$ canonical inducing directions $\bV_i = I$, through which we recover the full variational GP with derivatives as described in section \ref{sec:grad_svgp}. 

\paragraph{Complexity.} For a minibatch size $B$, when learning $p$ directional derivatives per inducing point, the matrices $\boK_{XZ}$ and $\boK_{ZZ}$ become $B \times M(p+1)$ and $M(p+1) \times M(p+1)$ respectively. As a result, the time complexity of variational GP inference with directional derivatives is $O(M^3p^3)$. When using $p$ directions per inducing point, this is computationally equivalent to running SVGP with $p+1$ times as many inducing points. To counteract the additional matrix size, one may use the whitened formulation of variational inference \cite{matthews2017scalable} for GPs when computing equation \eqref{eq:var_post_mean_and_cov} and use contour integral quadrature as in \cite{pleiss2020CIQ}.

\section{Experiments}
\label{sec:experiments}
In this section we compare the empirical performance of variational GPs with directional derivatives to the performance of variational GPs with derivatives, DSKI and DKIP on low dimensional regression problems, as well as compare to variational GPs without derivatives on high dimensional regression and Bayesian optimization (BO) tasks. All of our GP models use a constant prior and Gaussian kernel (or associated directional derivative kernel) and were accelerated through GPyTorch \cite{gardner2018gpytorch} on a single GPU. We also investigate the value of learning directional derivative information when derivative observations are not available through regression on common UCI-datasets.

\subsection{Synthetic functions}
\label{sec:synthetic}
In order to verify that variational GPs with directional derivatives perform well on basic learning tasks we perform a series of synthetic regression tasks. We consider low-dimensional regression with derivatives on test functions including Branin (2D), SixHumpCamel (2D), Styblinksi-Tang (2D) and Hartmann (6D) from \cite{surjanovic2013testfuns}, a modified 20D Welch test function \cite{ben2007welch} (Welch-m) \footnote{The Welch test function has intrinsically a 6D active space. We modified it to have a low-quality 6D active subspace and to show the limitation of GradSVGP and GradPPGPR.}, and a 5D sinusoid $f(x) = \sin(2\pi||x||^2)$ (Sin-5). We compare variational GPs without derivatives (SVGP, PPGPR) to variational GPs with derivatives (GradSVGP, GradPPGPR), exact GPs with derivatives (GradGP), non-variational GPs with derivatives (DSKI), and variational GPs with $p=2$ directional derivatives per inducing point (DSVGP2,DPPGPR2). 
Exact and variational GPs with derivatives are only tractable in low-dimensional settings due to the scalability issues mentioned in sections \ref{sec:related_work} and \ref{sec:grad_svgp}; therefore, to apply GradSVGP and GradPPGPR on the 20D Welch-m function, we first perform dimension reduction onto a low dimensional active subspace \cite{constantine2015active}, similar to \cite{eriksson2018scaling}. An active subspace of dimension $k$ is found by taking the first $k$ singular vectors of the matrix $P = \sum_{i=1}^N \nabla f(x_i)\nabla f(x_i)^T$, denoted $P_k$. The dimension-reduced data set is given by the triplets $\{(P_k^Tx_i,f(x_i),P_k^T\nabla f(x_i))\}_{i=1}^N$. To show the limitation of GradSVGP and GradPPGPR, we modified the Welch function to have a low-quality low-dimensional active subspace.

\begin{figure}[!h]
  \centering
  \begin{minipage}{\textwidth}
    \includegraphics[width=\textwidth]{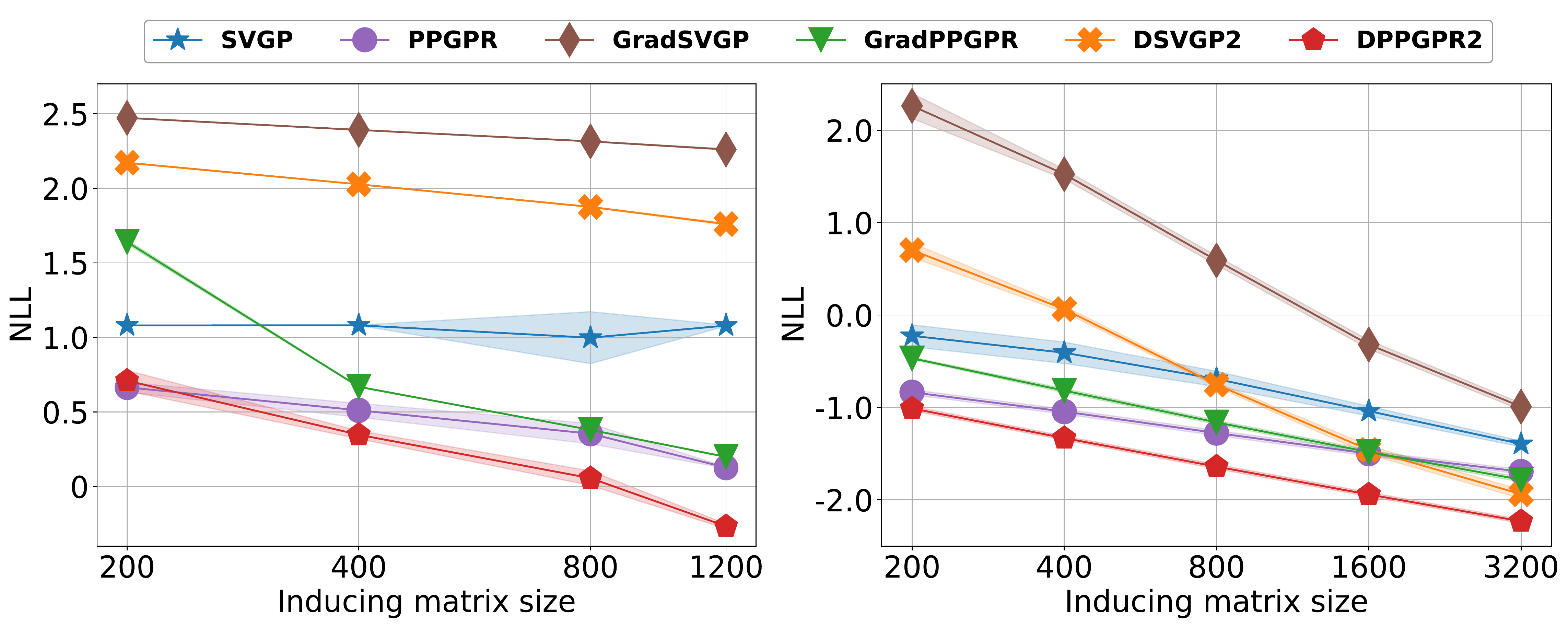}
    \end{minipage}
    \vspace{-10pt}
    \caption{Negative Log Likelihood (NLL) for the various GPs when using different inducing matrix sizes to regress on Sin-5 (Left) and Hartmann (Right). DPPGPR2 often outperforms other methods due to its ability incorporate derivative information while only modestly increasing the inducing matrix size. DSKI is removed because it does not have comparable matrix size.}
    \label{fig:synthetic}
\end{figure}

In this low-dimensional setting, we find that variational GPs with directional derivatives, DSVGP2 and DPPGPR2, perform comparably to the methods that incorporate full derivatives (DKSI, GradSVGP, GradPPGPR, GradGP); see Table \ref{table:synthetic}. In Figure \ref{fig:synthetic} we compare the negative log likelihood of each method as the inducing matrix size grows on the Sin-5 and Hartmann test functions. We find that DSVGP2 and DPPGPR2 often outperform other methods due to their ability incorporate derivative information while only modestly increasing the inducing matrix size.

\begin{table*}[!h]
\centering
\small
\tabcolsep 3pt
\begin{tabular}{rcccccccccccc}
\multicolumn{1}{c}{}  & \multicolumn{2}{c}{\textbf{Branin}}     & \multicolumn{2}{c}{\textbf{Camel}}      & \multicolumn{2}{c}{\textbf{StyTang}}   & \multicolumn{2}{c}{\textbf{Sin-5}}    & \multicolumn{2}{c}{\textbf{Hartmann}} & \multicolumn{2}{c}{\textbf{Welch-m}}   \\ \hline
\multicolumn{1}{l}{\textbf{}} & \begin{tabular}[c]{@{}c@{}}RMSE \\ (1e-3)\end{tabular} & NLL    & \begin{tabular}[c]{@{}c@{}}RMSE \\ (1e-3)\end{tabular} & NLL    & \begin{tabular}[c]{@{}c@{}}RMSE\\ (1e-3)\end{tabular} & NLL    & \begin{tabular}[c]{@{}c@{}}RMSE\\ (1e-1)\end{tabular} & NLL   & \begin{tabular}[c]{@{}c@{}}RMSE \\ (1e-1)\end{tabular} & NLL    & \begin{tabular}[c]{@{}c@{}}RMSE \\ (1e-2)\end{tabular} & NLL    \\ \hline
SVGP & 1.45   & -3.12 & 5.28   & -2.95  & 3.64  & -3.06  & 6.64  & 0.99  & 1.02   & -0.69  & 16.20  & -0.39  \\
PPGPR & 1.60   & -3.21  & 6.46   & -3.10  & 4.64  & -3.17  & 4.35  & 0.35  & 3.02   & -1.28  & 18.08  & -0.56  \\ \hline
\multicolumn{1}{c}{GradGP}    & 15.4   & -0.87  & 25.1   & -0.22  & 44.4  & -0.82  & \textbf{2.59} & \textbf{-.23} & \textbf{0.50}  & -0.74  & 16.3   & -0.38  \\
GradSVGP & 0.35   & -3.65  & 2.09  & \textbf{-3.62} & 1.00  & -3.65  & 4.85  & 2.31  & 2.08   & 0.59   & 18.94  & 42.82  \\
\multicolumn{1}{l}{GradPPGPR} & 0.67   & -3.32  & 23.1  & -3.14  & 2.91  & -3.30  & 4.83  & 0.37  & 3.95   & -1.16  & 18.92  & -0.25  \\ \hline
DSVGP2& \textbf{0.29}  & -3.10  & \textbf{1.82}  & -2.50 & \textbf{0.86} & -2.97  & 3.03  & 1.87  & 0.92   & -0.75 & \textbf{3.74}  & \textbf{-0.74} \\
DPPGPR2   & 0.47   & -3.32  & 8.43   & -3.24  & 1.75  & -3.31  & 4.30  & 0.05  & 2.69   & \textbf{-1.64} & 26.08  & -0.71  \\ \hline
DSKI  & 0.91   & \textbf{-4.47} & 3.85   & -3.00  & 1.59  & \textbf{-4.74} & N/A   & N/A   & N/A    & N/A    & N/A    & N/A   
\end{tabular}
\vspace{-6pt}
\caption{Regression results on Branin (2D), SixHumpCamel (2D), Styblinksi-Tang (2D), Sin-5 (5D), Hatrmann (6D) and Welch-m (20D), each with $10000$ training and $10000$ testing points. 
Following \cite{eriksson2018scaling}, we train GradGP on $10000/(D+1)$ points.
The inducing matrix size is 800 for all variational inducing point methods, while DSKI is trained on $800$ inducing points per dimension. 
}
\label{table:synthetic}
\end{table*}

\subsection{Implicit Surface Reconstruction}
\begin{figure}[!h]
\vspace{-5pt}
  \begin{minipage}{\textwidth}
  \centering
     \includegraphics[width=0.8\textwidth]{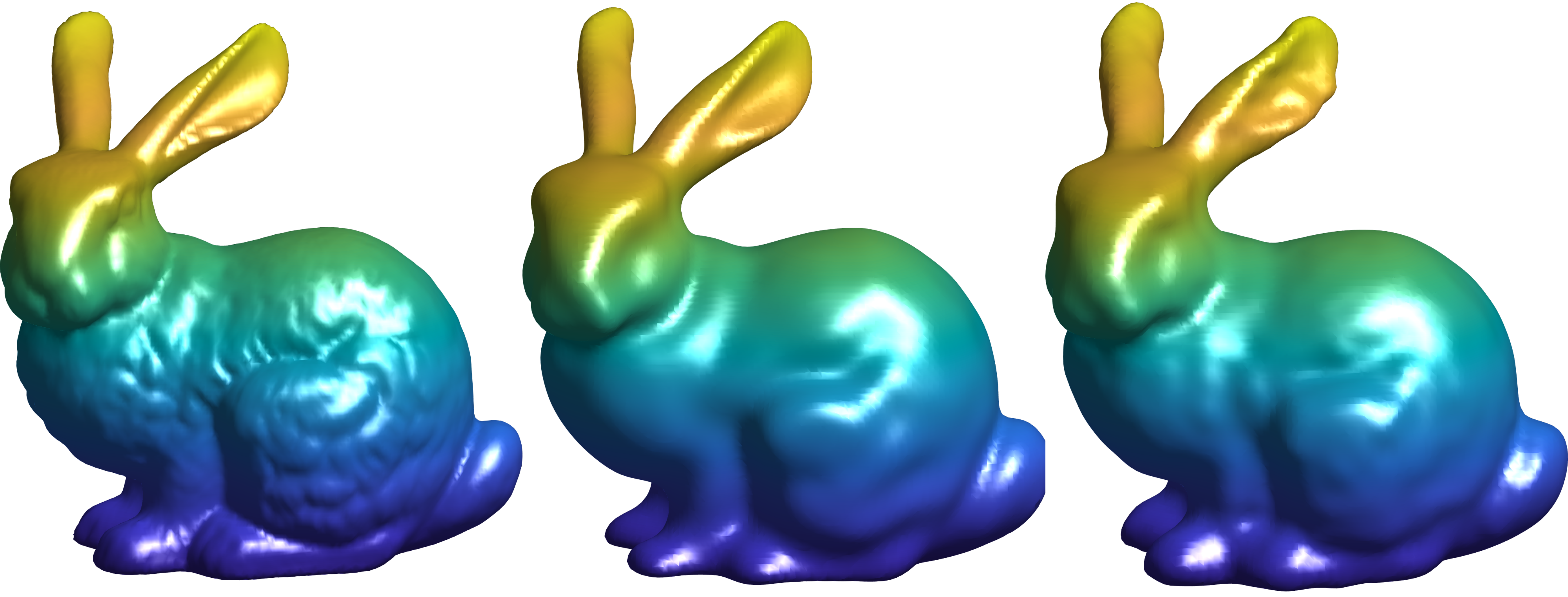}
    \end{minipage}
\vspace{-5pt}
  \caption{Surface reconstruction of the Stanford bunny: (Left) Original surface, (Middle) DSVGP with 800 inducing points and 3 directions, (Right) D-SKI with $30^3$ inducing grid points. }
  \label{fig:bunny}
\end{figure}

In order to further validate the fidelity of our method's derivative modeling, we consider the surface reconstruction task considered in \cite{eriksson2018scaling}. We compare to DSKI with the goal of achieving comparable performance, as DSKI is nearly exact for this problem. In \autoref{fig:bunny}, we reconstruct the Stanford Bunny by training DSVGP with $p=3$ inducing directions for $1200$ epochs and DSKI on 11606 noisy observations of 34818 locations and corresponding noise-free surface normals (gradients of the bunny level sets). DSVGP smoothly reconstructs the bunny and is comparable to DSKI.

\subsection{Training Graph Convolutional Neural Networks with Bayesian Optimization}
In this section, we demonstrate the full scalability of our approach by training the $D=4035$ parameters of a two layer graph convolutional neural network (GCN) \cite{Kipf2017gcn} on the node classification task of the Pubmed citation dataset \cite{sen2008pubmed} using Bayesian optimization. The Bayesian optimization setting compounds the need for scalability, as the GP model must be retrained after each batch of data is acquired. For example, in the last $500$ of $2500$ optimization iterations with a batch size of 10, a GP must be fit $50$ times to datasets with $N(D+1) \approx2500(4035+1) >10^6$ combined function and partial derivative labels. Any one of these datasets would be intractable to existing methods for training GPs with gradient observations. 

\begin{figure}[!h]
  \centering
    \begin{minipage}{\textwidth}
     \includegraphics[width=\textwidth]{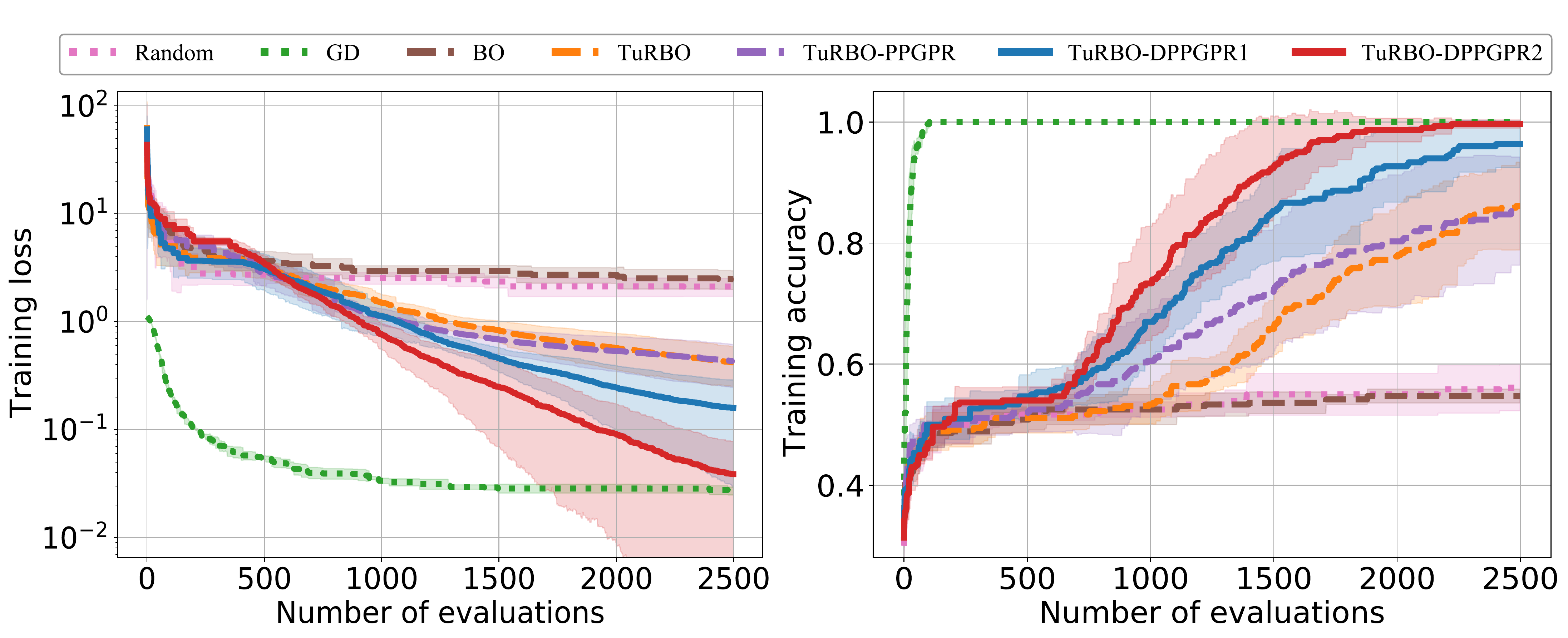}
    \end{minipage}
    \vspace{-10pt}
  \caption{GCN training on the Pubmed dataset: (Left) training loss and (Right) training accuracy. Averaged over 6 trials for all optimizers. \label{figure: GCN}}
\end{figure}

For this experiment, we make no effort to modify the Bayesian optimization routine itself to account for the derivative information (e.g., as in \cite{wu2017BOGradients}), as this would confound the performance improvements achieved by higher fidelity modelling by incorporating derivative information. Instead, we focus only on swapping out the underlying Gaussian process model. We consider TuRBO \cite{eriksson2019turbo} as a base Bayesian optimization algorithm which we run with an exact GP, PPGPR, DPPGPR1 and DPPGPR2 surrogate models. We additionally include traditional BO with the Lower Confidence Bound (LCB) \cite{srinivas2012UCB} acquisition function, gradient descent and random search. All algorithms were initialized with 400 random evaluations, and the TuRBO variants were run with a batch size of 20 and retrained over $150$ steps. Figure \ref{figure: GCN} summarizes results averaged over 6 trials. We observe that TuRBO with DPPGPR significantly outperform traditional BO and other TuRBO variants. While all Bayesian optimization methods under-perform compared to standard gradient descent, we conjecture that this performance gap could be narrowed by incorporating the gradient information into the Bayesian optimization algorithm itself.

\vspace{-10pt}

\subsection{Stellarator Regression}
In this experiment we show the capacity of variational GPs with directional derivatives to extend GP regression with derivatives to massive datasets in a high dimensional settings. We perform regression on $N=500000$ function and gradient observations gathered from a $D=45$ dimensional optimization objective function through the FOCUS code \cite{zhu2017FOCUS}: a code for evaluating the quality of magnetic coils for a Stellarator, a magnetic confinement based fusion device for generating renewable energy \cite{imbert2019stellarator}. The dataset is available upon request.

\begin{figure}[!h]
  \centering
  \begin{minipage}{\textwidth}
     \includegraphics[width=\textwidth]{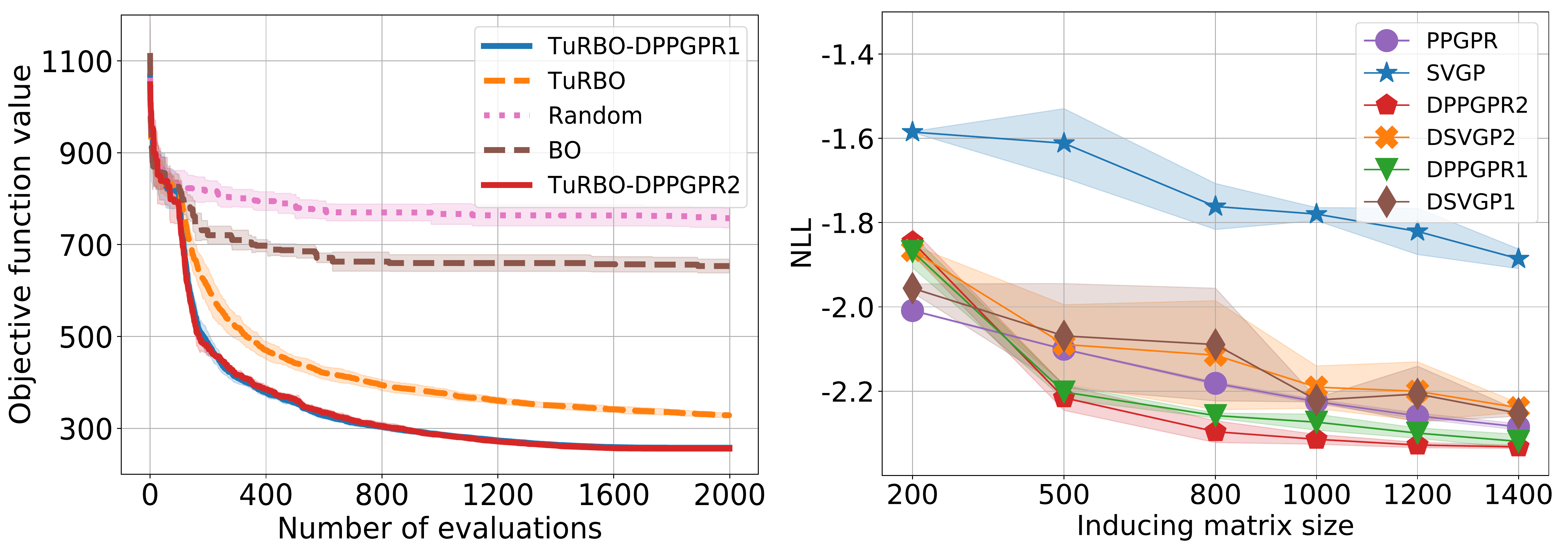}
    \end{minipage}
  \vspace{-10pt}
  \caption{\textbf{Rover} (Left) and \textbf{Stellarator Regression} (Right). Negative log likelihood of GP variants as the inducing matrix size increases for the $D=45, N=500000$ Stellarator Regression experiment. Rover (Left) shows the value of the objective function over the course of optimization.}
  \label{fig:rover_stellarator}
\end{figure}

We compare variational GPs with directional derivatives using $p=1,2$ directions (DSVGP1,DPPGPR1,DSVGP2,DPPGPR2) to variational GPs without directional derivatives (SVGP,PPGPR). While $D$ is too large to use variational or exact GPs with derivatives, we can apply variational GPs with derivatives to a projection of the data set onto a low-dimensional active subspace as in section \ref{sec:synthetic}. Variational GPs with derivatives trained on reduced datasets of dimension two and three performed poorly compared to all other methods tested. The results of this experiment are shown in Figure \ref{fig:rover_stellarator}: variational GPs with directional derivatives significantly enhance regression performance. Even the inclusion of one directional derivative is enough the enhance the predictive capabilities of the regressor. The experiments were averaged over 5 trials, using a Adam with a Multi Step learning rate schedule and 1000 epochs.

\subsection{Rover Trajectory Planning}
The rover trajectory planning problem \cite{eriksson2019turbo,wang2018batched} is a $D=200$ dimensional optimization problem with gradients. This experiment validates the use of variational GPs with directional derivatives in Bayesian optimization by leveraging derivative information in a setting where no other method can. We solve a variant of the rover problem in which the goal is find an open-loop controller that minimizes the energy of guiding a rover through a series of waypoints in the $xy$-plane. The rover trajectory is integrated over $100$ steps at which forces in $x$ and $y$ directions are applied to the rover, making a total of $D=200$ decision variables. We compare the performance of TuRBO, TuRBO with DPPGPR using $p=1$, TuRBO with DPPGPR using $p=2$, traditional Bayesian optimization with the LCB acquisition function, and random search. All algorithms were initialized with a $100$ point experimental design, and the TuRBO variants were run with a batch size of $5$, and retrained over $300$ steps. The results, averaged over 5 trials, are summarized in figure \ref{fig:rover_stellarator}. We observe that the TuRBO variants that leverage derivative information outperform the other algorithms almost immediately.

\subsection{UCI Regression}
Increasing the number of inducing points for Gaussian process models often results in diminishing returns on final model performance, with $500 \leq M \leq 2000$ often proving sufficient \cite{hensman2013gaussian,salimbeni2018orthogonally,jankowiak2020parametric,pleiss2020CIQ}. This saturation is likely due in part to the ability of sparse Gaussian processes to represent variation in the data, but may also be due to increasingly challenging optimization dynamics as more inducing points are added.

One hypothesis worth exploring is that, in some cases, it may be beneficial to augment a smaller set of inducing points with additional descriptive variables rather than to simply increase the number of inducing points. To that end, we test our method on a number of UCI benchmark regression datasets \emph{for which no derivative information is available}. In this setting, running DSVGP1 or DPPGPR1 involves maintaining a model with inducing directional derivatives as normal, but minibatches of data always correspond to labels. In other words, rows of the matrix $\boK_{XZ}$ never correspond to partial derivative observations because there are none.

We test our methodology on a number of UCI datasets \cite{DuaUCI}: Protein (D=9, N=45730), Elevators (D=18 N=16599), Kin40k (D=8, N=40000), Sydney (D=32, N=72000), Kegg-Directed (D=20, N=53414). We use an 80-20 train-test split for all experiments, with $M=500$ inducing points and $p=1$ direction for DSVGP1 and DPPGPR1, and $M=1000$ inducing points for SVGP and PPGPR to ensures that the sizes of the inducing matrices are the same. We train for $300$ epochs with a mini-batch size of $500$ using Adam \cite{kingma2014adam} with a learning rate of $0.01$. Interestingly, the results in \autoref{tab:uci} show learning derivative information can improve prediction performance. 
\begin{table*}[!h]
\centering
\small
\tabcolsep 3pt
\begin{tabular}{rcccccccccc}
\multicolumn{1}{l}{} & \multicolumn{2}{c}{Elevators}  & \multicolumn{2}{c}{kin40k} & \multicolumn{2}{c}{Sydney} & \multicolumn{2}{c}{Protein}  & \multicolumn{2}{c}{Kegg-directed}\\\hline
\multicolumn{1}{l}{} & \multicolumn{1}{c}{\begin{tabular}[c]{@{}c@{}}MSE\\ (1e-1)\end{tabular}} & \multicolumn{1}{c}{\begin{tabular}[c]{@{}c@{}}NLL\\ (1e-1)\end{tabular}} & \multicolumn{1}{c}{\begin{tabular}[c]{@{}c@{}}MSE\\ (1e-2)\end{tabular}} & \multicolumn{1}{c}{\begin{tabular}[c]{@{}c@{}}NLL\\ (1e-1)\end{tabular}} & \multicolumn{1}{c}{\begin{tabular}[c]{@{}c@{}}MSE\\ (1e-1)\end{tabular}} & \multicolumn{1}{c}{\begin{tabular}[c]{@{}c@{}}NLL\\ (1e-1)\end{tabular}} & \multicolumn{1}{c}{\begin{tabular}[c]{@{}c@{}}MSE\\ (1e-1)\end{tabular}} & \multicolumn{1}{c}{NLL} & \multicolumn{1}{c}{\begin{tabular}[c]{@{}c@{}}MSE\\ (1e-3)\end{tabular}} & \multicolumn{1}{c}{NLL} \\ \hline
SVGP & 1.48& 4.66 & 4.22 & -1.09& 1.51 & 4.61 & 5.89 & 1.15& 9.37 & -0.92 \\
PPGPR& 1.63 & 3.76 & 7.81 & -6.71& 1.40 & 1.44   & 5.97   & 1.03& 9.64 & -1.25 \\ \hline
DSVGP1 & \textbf{1.41} & 4.39   & \textbf{2.67} & -3.61  & 1.29   & 3.75 & \textbf{5.51} & 1.12& 8.94  & -0.95   \\
DPPGPR1 & 1.56   & \textbf{3.75} & 6.22   & \textbf{-8.60}& \textbf{1.28} & \textbf{1.08} & 5.76   & \textbf{1.01}   & \textbf{8.70} & \textbf{-1.38} 
\end{tabular}
\caption{Variational GPs with no derivatives (SVGP,PPGPR) and Variational GPs with $p=1$ directions (DSVGP1,DPPGPR1) on UCI benchmark regression datasets for which no derivative information is available.}
\label{tab:uci}
\end{table*}
\section{Discussion}
\label{sec:discussion}
Augmenting GPs with derivative information can significantly improve their predicitive capabilities; however, with the benefits comes a significant $O(N^3D^3)$ cost of training and inference. We introduce a novel method for achieving fully scalable --- scalable in $N$ and $D$ --- GPs with derivative information by leveraging stochastic variational approximations. The resulting model reduces the cost of training GPs with derivatives to $O(M^3p^3)$ time per iteration of training, where $M \ll N$ and $p \ll D$. A practical limitation of our method is that $M,p$ must be small enough for fast computations, which is not a reasonable assumption is very high dimensional problems. Through a series of synthetic experiments and a surface reconstruction task, we demonstrate the quality of our approximate model in low dimensional settings. Furthermore, we demonstrate the full scalability of our model through training a graph convolutional neural network using Bayesian optimization, in addition to performing regression on a large scale Stellarator fusion dataset with derivatives. Lastly, we show that our methods can even have benefit in the regression setting when no derivative information is available in the training set, by including a new avenue to encode information. While this last result is a surprising benefit of GPs with derivatives, it is not well understood and is thus a good direction for future study. 
While our method may make GPs more accessible to practitioners and researchers for calibrating uncertainty estimates, the fundamental assumption that the data is drawn from a GP may flawed, leading to poor uncertainty estimates and a lack of robustness altogether. Researchers and practitioners should take care to understand the reliability of the GP model in their setting rather than relying faithfully on a black-box approach.

\section{Acknowledgements}
We acknowledge support from Simons Foundation Collaboration on Hidden Symmetries and Fusion Energy and the National Science Foundation NSF CCF-1934985, and NSF DMS-1645643.

{
\small
\bibliographystyle{plain}
\bibliography{./ref.bib}

}

\end{document}